\documentclass[11pt,a4paper]{article}
\usepackage{authblk}
\usepackage[hyperref]{acl2018}
\usepackage{times}
\usepackage{latexsym}
\usepackage{times,amsmath,amssymb,subcaption}
\usepackage{latexsym,graphicx,color,array,authblk}
\usepackage{multirow}
\usepackage{url,hyperref}
\hypersetup{
  urlcolor = blue
}
\DeclareMathOperator{\mW}{\mathbf{W}}
\DeclareMathOperator{\vw}{\mathbf{w}}
\DeclareMathOperator{\fe}{\mathbf{\Phi}}

\DeclareMathOperator{\R}{\mathbb{R}}

\def\da{DA}
\newcolumntype{M}{>{\centering\arraybackslash}m{4.2cm}}
\newcolumntype{P}{>{\centering\arraybackslash}m{1.4cm}}
\aclfinalcopy 

\title{Domain Adapted Word Embeddings for Improved Sentiment Classification}

\author[]{Prathusha K Sarma}
\author[]{Yingyu Liang}
\author[]{William A Sethares}

\affil[]{University of Wisconsin-Madison \\
         {\tt \{kameswarasar,sethares\}@wisc.edu},\\
         \tt yliang@cs.wisc.edu}

\date{}

\begin{document}
\maketitle
\begin{abstract}
{\em Generic} word embeddings are trained on large-scale generic corpora; {\em Domain Specific} (DS) word embeddings are trained only on data from a domain of interest. This paper proposes a method to combine the breadth of generic embeddings with the specificity of domain specific embeddings. The resulting embeddings, called {\em Domain Adapted} (\da) word embeddings, are formed by aligning corresponding word vectors using Canonical Correlation Analysis (CCA) or the related nonlinear Kernel CCA. 
Evaluation results on sentiment classification tasks show that the DA embeddings substantially outperform both generic and DS embeddings when used as input features to standard or state-of-the-art sentence encoding algorithms for classification. 
\end{abstract}

\section{Introduction}
Generic word embeddings such as Glove and word2vec~\cite{pennington2014glove,mikolov2013distributed} which are pre-trained on large sets of raw text, have demonstrated remarkable success when used as features to a supervised learner in various applications such as the sentiment classification of text documents. There are, however, many applications with domain specific vocabularies and relatively small amounts of data. The performance of generic word embedding in such applications is limited, since word embeddings pre-trained on generic corpora do not capture domain specific semantics/knowledge, while embeddings learned on small data sets are of low quality. 

A concrete example of a small-sized domain specific corpus is the Substances User Disorders (SUDs) data set~\cite{quanbeck2014mobile,litvin2013computer}, which contains messages on discussion forums for people with substance addictions. These forums are part of a mobile health intervention treatment that encourages participants to engage in sobriety-related discussions. The goal of such treatments is to analyze content of participant's digital media content and provide human intervention via machine learning algorithms. This data is both domain specific and limited in size.
Other examples include customer support tickets reporting issues with taxi-cab services, product reviews, reviews of restaurants and movies, discussions by special interest groups and political surveys. In general they are common in domains where words have different sentiment from what they would have elsewhere.

Such data sets present significant challenges for word embedding learning algorithms.
First, words in data on specific topics have a different distribution than words from generic corpora. Hence using generic word embeddings obtained from algorithms trained on a corpus such as Wikipedia, may introduce considerable errors in performance metrics on specific downstream tasks such as sentiment classification.  
For example, in SUDs, discussions are focused on topics related to recovery and addiction; 
the sentiment behind the word `party' may be very different in a dating context than in a substance abuse context.
Thus domain specific vocabularies and word semantics may be a problem for pre-trained sentiment classification models \cite{blitzer2007biographies}. 
Second, there is insufficient data to completely retrain a new set of word embeddings. The SUD data set consists of a few hundred people and only a fraction of these are active~\cite{firth2017can}, \cite{naslund2015emerging}. This results in a small data set of text messages available for analysis. Furthermore, content is generated spontaneously on a day to day basis, and language use is informal and unstructured.  
Fine-tuning the generic word embedding also leads to noisy outputs due to the highly non-convex training objective and the small amount of data. Since such data sets are common, a simple and effective method to adapt word embedding approaches is highly valuable. While existing work~\cite{yin2016learning},~\cite{luo2014pre},~\cite{mehrkanoon2017regularized},~\cite{anoop2015utility},~\cite{blitzer2011domain} combines word embeddings from different algorithms to improve upon intrinsic tasks such as similarities, analogies etc, there does not exist a concrete method to combine multiple embeddings to perform domain adaptation or improve on extrinsic tasks.

This paper proposes a method for obtaining high quality word embeddings that capture domain specific semantics and are suitable for tasks on the specific domain. The new Domain Adapted (\da) embeddings are obtained by combining generic embeddings and Domain Specific (DS) embeddings via CCA/KCCA. Generic embeddings are trained on large corpora and do not capture domain specific semantics, while DS embeddings are obtained from the domain specific data set via algorithms such as Latent Semantic Analysis (LSA) or other embedding methods. The two sets of embeddings are combined using a linear CCA~\cite{hotelling1936relations} or a nonlinear kernel CCA (KCCA)~\cite{hardoon2004canonical}. They are projected along the directions of maximum correlation, and a new (\da) embedding is formed by averaging the projections of the generic embeddings and DS embeddings. The \da~embeddings are then evaluated in a sentiment classification setting. Empirically, it is shown that the CCA/KCCA combined \da~embeddings improve substantially over the generic embeddings, DS embeddings and a concatenation-SVD (concSVD) based baseline.

The remainder of this paper is organized as follows. Section~\ref{ccafor} briefly introduces the CCA/KCCA and details the procedure used to obtain the \da~embeddings. Section~\ref{exps} describes the experimental set up. Section~\ref{results} discusses the results from sentiment classification tasks on benchmark data sets using standard classification as well as using a sophisticated neural network based sentence encoding algorithm. Section~\ref{conclusion} concludes this work.   

\section{Domain Adapted Word Embeddings}\label{ccafor}
Training word embeddings directly on small data sets leads to noisy outputs while embeddings from generic corpora fail to capture specific local meanings within the domain. Here we combine DS and generic embeddings using CCA KCCA, which projects corresponding word vectors along the directions of maximum correlation.

Let $\mW_{DS} \in \R^{|V_{DS}| \times d_{1}}$ be the matrix whose columns are the domain specific word embeddings (obtained by, e.g., running the LSA algorithm on the domain specific data set), where $V_{DS}$ is its vocabulary and $d_1$ is the dimension of the embeddings. 
Similarly, let $\mW_{G} \in \R^{|V_{G}| \times d_{2}}$ be the matrix of generic word embeddings (obtained by, e.g., running the GloVe algorithm on the Common Crawl data), where $V_{G}$ is the vocabulary and $d_2$ is the dimension of the embeddings. Let $V_\cap = V_{DS} \cap V_{G}$. Let $\vw_{i,DS}$ be the domain specific embedding of the word $i \in V_\cap$, and  $\vw_{i,G}$ be its generic embedding.
For one dimensional CCA, let $\phi_{DS}$ and $\phi_{G}$ be the projection directions of $\vw_{i,DS}$ and $\vw_{i,G}$ respectively. Then the projected values are,

\begin{align}
\nonumber
\bar{w}_{i,DS}& = \vw_{i,DS}\phi_{DS}\\
\bar{w}_{i,G}& = \vw_{i,G}\phi_{G} .
\end{align}
CCA maximizes the correlation between $\bar{w}_{i,DS}$ and $\bar{w}_{i,G}$ to obtain $\phi_{DS}$ and $\phi_{G}$ such that
\begin{align}\label{lincca}
\rho(\phi_{DS},\phi_{G}) = \max_{\phi_{DS},\phi_{G}}\frac{\mathbb{E}[\langle \bar{w}_{i,DS},\bar{w}_{i,G} \rangle]}{\sqrt{\mathbb{E}[\bar{w}_{i,DS}^{2}]\mathbb{E}[\bar{w}_{i,G}^{2}]}} 
\end{align}
where $\rho$ is the correlation between the projected word embeddings and $\mathbb{E}$ is the expectation over all words $i \in V_\cap$.

The $d$-dimensional CCA with $d>1$ can be defined recursively. Suppose the first $d-1$ pairs of canonical variables are defined. Then the $d^{th}$ pair is defined by seeking vectors maximizing the same correlation function subject to the constraint that they be uncorrelated with the first $d-1$ pairs.
Equivalently, matrices of projection vectors $\fe_{DS} \in \R^{d_{1} \times d}$ and $\fe_{G} \in \R^{d_{2} \times d}$ are obtained for all vectors in $\mW_{DS}$ and $\mW_{G}$ where $d \leq \min{\{d_{1},d_{2}\}}$. Embeddings obtained by $\bar{\vw}_{i,DS} = \vw_{i,DS}\fe_{DS}$ and $\bar{\vw}_{i,G} = \vw_{i,G}\fe_{G}$ are projections along the directions of maximum correlation.

The final domain adapted embedding for word $i$ is given by 
$
\hat{\vw}_{i,\da} = \alpha\bar{\vw}_{i,DS}+\beta\bar{\vw}_{i,G},
$ 
where the parameters $\alpha$ and $\beta$ can be obtained by solving the following optimization, 
\begin{multline}\label{opteq}
\min_{\alpha,\beta} \|\bar{\vw}_{i,DS} - (\alpha\bar{\vw}_{i,DS}+\beta\bar{\vw}_{i,G})\|_{2}^{2}+\\
\|\bar{\vw}_{i,G} -(\alpha\bar{\vw}_{i,DS}+\beta\bar{\vw}_{i,G})\|_{2}^{2}.
\end{multline}
Solving \eqref{opteq} gives a weighted combination with $\alpha = \beta = \frac{1}{2}$, i.e., the new vector is equal to the average of the two projections:
\begin{align}
\hat{\vw}_{i,\da} = \frac{1}{2}\bar{\vw}_{i,DS}+\frac{1}{2}\bar{\vw}_{i,G}.
\end{align} 

Because of its linear structure, the CCA in \eqref{lincca} may not always capture the best relationships between the two matrices. To account for nonlinearities, a kernel function, which implicitly maps the data into a high dimensional feature space, can be applied. For example, given a vector $\vw \in \R^{d}$, a kernel function $K$ is written in the form of a feature map $\varphi$ defined by $\varphi: \vw = (\vw_{1},\hdots,\vw_{d}) \mapsto \varphi(\vw) = (\varphi_{1}(\vw),\hdots,\varphi_{m}(\vw)) (d < m)$ such that given $\vw_{a}$ and $\vw_{b}$
\begin{align}
\nonumber
K(\vw_{a},\vw_{b}) = \langle \varphi(\vw_{a}),\varphi(\vw_{b})\rangle .
\end{align}
In kernel CCA, data is first projected onto a high dimensional feature space before performing CCA. In this work the kernel function used is a Gaussian kernel, i.e.,
\begin{align}
\nonumber
K(\vw_{a},\vw_{b}) = \text{exp} \Big(-\frac{||\vw_{a}-\vw_{b}||^{2}}{2\sigma^{2}}\Big) .
\end{align}
The implementation of kernel CCA follows the standard algorithm described in several texts such as \cite{hardoon2004canonical}; see reference for details.
\begin{table}[!h]
\centering
\resizebox{\linewidth}{!}{
\begin{tabular}{|c|@{}P@{}|@{}M@{}|c|c|c|}
\hline
Data Set& &Embedding & Avg Precision & Avg F-score & Avg AUC\\
\hline
Yelp&\begin{tabular}{c}
     \\
     \\
     \\
     \\
     \\
     $\mW_{\da}$\\
     \\
     \\
     \\
     \\
     \\
     \\
     $\mW_{G}$\\
     \\
     \\
     $\mW_{DS}$\\
     \end{tabular}
     &\begin{tabular}{c}
     KCCA(Glv, LSA)\\
     CCA(Glv, LSA)\\
     KCCA(w2v, LSA)\\
     CCA(w2v, LSA)\\
     \textbf{KCCA(GlvCC, LSA)}\\
     CCA(GlvCC, LSA)\\
     KCCA(w2v, DSw2v)\\
     CCA(w2v, DSw2v)\\
     concSVD(Glv, LSA)\\
     concSVD(w2v, LSA)\\
     concSVD(GlvCC, LSA)\\
     \hline
     GloVe\\
     GloVe-CC\\
     word2vec\\
     \hline
     LSA \\
     word2vec\\
     \end{tabular}&\begin{tabular}{c}
     85.36$\pm$ 2.8\\
     83.69$\pm$ 4.7\\
     87.45$\pm$ 1.2\\
     84.52$\pm$ 2.3\\
     \textbf{88.11$\pm$ 3.0}\\
     83.69$\pm$ 3.5\\
     78.09$\pm$ 1.7\\
     86.22$\pm$ 3.5\\
     80.14$\pm$ 2.6\\
     85.11$\pm$ 2.3\\
     84.20$\pm$ 3.7\\
     77.13$\pm$ 4.2\\
     82.10$\pm$ 3.5\\
     82.80$\pm$ 3.5\\
     75.36$\pm$ 5.4\\
     73.08$\pm$ 2.2
     \end{tabular}&\begin{tabular}{c}
     81.89$\pm$2.8\\
     79.48$\pm$2.4 \\
     83.36$\pm$1.2\\
     80.02$\pm$2.6\\
     \textbf{85.35$\pm$2.7}\\
     78.99$\pm$4.2\\
     76.04$\pm$1.7\\
     84.35$\pm$2.4\\
     78.50$\pm$3.0\\
     83.51$\pm$2.2\\
     80.39$\pm$3.7\\
     72.32$\pm$7.9\\
     76.74$\pm$3.4\\
     78.28$\pm$3.5\\
     71.17$\pm$4.3\\
     70.97$\pm$2.4
     \end{tabular}&\begin{tabular}{c}
     82.57$\pm$1.3\\
     80.33$\pm$2.9\\
     84.10$\pm$0.9\\
     81.04$\pm$2.1\\
     \textbf{85.80$\pm$2.4}\\
     80.03$\pm$3.7\\
     76.66$\pm$1.5\\
     84.65$\pm$2.2\\
     78.92$\pm$2.7\\
     83.80$\pm$2.0\\
     80.83$\pm$3.9\\
     74.17$\pm$5.0\\
     78.17$\pm$2.7\\
     79.35$\pm$3.1\\
     72.57$\pm$4.3\\
     71.76$\pm$2.1
     \end{tabular}\\
\hline
Amazon&\begin{tabular}{c}
     \\
     \\
     \\
     \\
     \\
     $\mW_{\da}$\\
     \\
     \\
     \\
     \\
     \\
     \\
     $\mW_{G}$\\
     \\
     \\
     $\mW_{DS}$\\
     \end{tabular}
     &\begin{tabular}{c}
     KCCA(Glv, LSA)\\
     CCA(Glv, LSA)\\
     KCCA(w2v, LSA)\\
     CCA(w2v, LSA)\\
     \textbf{KCCA(GlvCC, LSA)}\\
     CCA(GlvCC, LSA)\\
     KCCA(w2v, DSw2v)\\
     CCA(w2v, DSw2v)\\
     concSVD(Glv, LSA)\\
     \textbf{concSVD(w2v, LSA)}\\
     concSVD(GlvCC, LSA)\\
     \hline
     GloVe\\
     GloVe-CC\\
     word2vec\\
     \hline
     LSA \\
     word2vec\\
     \end{tabular}&\begin{tabular}{c}
     86.30$\pm$1.9\\
     84.68$\pm$2.4\\
     87.09$\pm$1.8\\
     84.80$\pm$1.5\\
     \textbf{89.73$\pm$2.4}\\
     85.67$\pm$2.3\\
     85.68$\pm$3.2\\
     83.50$\pm$3.4\\
     82.36$\pm$2.0\\
     87.28$\pm$2.9\\
     84.93$\pm$1.6\\
     81.58$\pm$2.5\\
     79.91$\pm$2.7\\
     84.55$\pm$1.9\\
     82.65$\pm$4.4\\
     74.20$\pm$5.8
     \end{tabular}&\begin{tabular}{c}
     83.00$\pm$2.9\\
     82.27$\pm$2.2\\
     82.63$\pm$2.6\\
     81.42$\pm$1.9\\
     85.47$\pm$2.4\\
     83.83$\pm$2.3\\     
     81.23$\pm$3.2\\
     81.31$\pm$4.0\\
     81.30$\pm$3.5\\
     \textbf{86.17$\pm$2.5}\\
     77.81$\pm$2.3\\
     77.62$\pm$2.7\\
     81.63$\pm$2.8\\
     80.52$\pm$2.5\\
     73.92$\pm$3.8\\
     72.49$\pm$5.0
     \end{tabular}&\begin{tabular}{c}
     83.39$\pm$3.2\\
     82.78$\pm$1.7\\
     83.50$\pm$2.0\\
     82.12$\pm$1.3\\
     85.56$\pm$2.6\\
     84.21$\pm$2.1\\
     82.20$\pm$2.9\\
     81.86$\pm$3.7\\
     81.51$\pm$2.5\\
     \textbf{86.42$\pm$2.0}\\
     79.52$\pm$1.7\\
     78.72$\pm$2.7\\
     81.46$\pm$2.6\\
     81.45$\pm$2.0\\
     76.40$\pm$3.2\\
     73.11$\pm$4.8
     \end{tabular}\\
\hline
IMDB&\begin{tabular}{c}
     \\
     \\
     \\
     \\
     \\
     \da\\
     \\
     \\
     \\
     \\
     \\
     \\
     $\mW_{G}$\\
     \\
     \\
     $\mW_{DS}$\\
     \end{tabular}
     &\begin{tabular}{c}
     KCCA(Glv, LSA)\\
     CCA(Glv, LSA)\\
     \textbf{KCCA(w2v, LSA)}\\
     CCA(w2v, LSA)\\
     KCCA(GlvCC, LSA)\\
     CCA(GlvCC, LSA)\\
     KCCA(w2v, DSw2v)\\
     CCA(w2v, DSw2v)\\
     concSVD(Glv, LSA)\\
     concSVD(w2v, LSA)\\
     concSVD(GlvCC, LSA)\\
     \hline
     GloVe\\
     GloVe-CC\\
     word2vec\\
     \hline
     LSA \\
     word2vec\\
     \end{tabular}&\begin{tabular}{c}
     73.84$\pm$1.3\\
     73.35$\pm$2.0\\
     \textbf{82.36$\pm$4.4}\\
     80.66$\pm$4.5\\
     54.50$\pm$2.5\\
     54.08$\pm$2.0\\     
     60.65$\pm$3.5\\
     58.47$\pm$2.7\\
     73.25$\pm$3.7\\
     53.87$\pm$2.2\\
     78.28$\pm$3.2\\
     64.44$\pm$2.6\\
     50.53$\pm$1.8\\
     78.92$\pm$3.7\\
     67.92$\pm$1.7\\
     56.87$\pm$3.6
     \end{tabular}&\begin{tabular}{c}
     73.07$\pm$3.6\\
     73.00$\pm$3.2\\
     \textbf{78.95$\pm$2.7}\\
     75.95$\pm$4.5\\
     54.42$\pm$2.9\\ 
     53.03$\pm$3.5\\    
     58.95$\pm$3.2\\
     57.62$\pm$3.0\\
     74.55$\pm$3.2\\
     51.77$\pm$5.8\\
     77.67$\pm$3.7\\
     65.18$\pm$3.5\\
     62.39$\pm$3.5\\
     74.88$\pm$3.1\\
     69.79$\pm$5.3\\
     56.04$\pm$3.1
     \end{tabular}&\begin{tabular}{c}
     73.17$\pm$2.4\\
     73.06$\pm$2.0\\
     \textbf{79.66$\pm$2.6}\\
     77.23$\pm$3.8\\
     53.91$\pm$2.0\\
     54.90$\pm$2.1\\     
     58.95$\pm$3.7\\
     58.03$\pm$3.9\\
     73.02$\pm$4.7\\
     53.54$\pm$1.9\\
     74.55$\pm$2.9\\
     64.62$\pm$2.6\\
     49.96$\pm$2.3\\
     75.60$\pm$2.4\\
     69.71$\pm$3.8\\
     59.53$\pm$8.9
     \end{tabular}\\
\hline
A-CHESS
&\begin{tabular}{c}
     \\
     \\
     \\
     \\
     \\
     \da\\
     \\
     \\
     \\
     \\
     \\
     \\
     $\mW_{G}$\\
     \\
     \\
     $\mW_{DS}$\\
     \end{tabular}
&\begin{tabular}{c}
     \textbf{KCCA(Glv, LSA)}\\
     CCA(Glv, LSA)\\
     \textbf{KCCA(w2v, LSA)}\\
     CCA(w2v, LSA)\\
     KCCA(GlvCC, LSA)\\
     CCA(GlvCC, LSA)\\
     KCCA(w2v, DSw2v)\\
     CCA(w2v, DSw2v)\\
     concSVD(Glv, LSA)\\
     concSVD(w2v, LSA)\\
     concSVD(GlvCC, LSA)\\
     \hline
     GloVe\\
     \textbf{GloVe-CC}\\
     word2vec\\
     \hline
     LSA \\
     word2vec\\
     \end{tabular}&\begin{tabular}{c}
     32.07$\pm$1.3\\
     32.70$\pm$1.5\\
     33.45$\pm$1.3\\
     33.06$\pm$3.2\\
     36.38$\pm$1.2\\
     32.11$\pm$2.9\\
     25.59$\pm$1.2\\
     24.88$\pm$1.4\\
     27.27$\pm$2.9\\
     29.84$\pm$2.3\\
     28.09$\pm$1.9\\
     30.82$\pm$2.0\\
     \textbf{38.13$\pm$0.8}\\
     32.67$\pm$2.9\\
     27.42$\pm$1.6\\
     24.48$\pm$0.8
     \end{tabular}&\begin{tabular}{c}
     39.32$\pm$2.5\\
     35.48$\pm$4.2\\
     \textbf{39.81$\pm$1.0}\\
     34.02$\pm$1.1\\
     34.71$\pm$4.8\\
     36.85$\pm$4.4\\
     28.27$\pm$3.1\\
     29.17$\pm$3.1\\
     34.45$\pm$3.0\\
     36.32$\pm$3.3\\
     35.06$\pm$1.4\\
     33.67$\pm$3.4\\
     27.45$\pm$3.1\\
     31.72$\pm$1.6\\
     34.38$\pm$2.3\\
     27.97$\pm$3.7
     \end{tabular}&\begin{tabular}{c}
     \textbf{65.96$\pm$1.3}\\
     62.15$\pm$2.9\\
     65.92$\pm$0.6\\
     60.91$\pm$0.9\\
     61.36$\pm$2.6\\
     62.99$\pm$3.1\\
     57.25$\pm$1.7\\
     57.76$\pm$2.0\\
     61.59$\pm$2.3\\
     62.94$\pm$1.1\\
     62.13$\pm$2.6\\
     60.80$\pm$2.3\\
     57.49$\pm$1.2\\
     59.64$\pm$0.5\\
     61.56$\pm$1.9\\
     57.08$\pm$2.5
     \end{tabular}\\
\hline
\end{tabular}
}
\caption{This table shows results from the classification task using sentence embeddings obtained from weighted averaging of word embeddings. Metrics reported are average Precision, F-score and AUC and the corresponding standard deviations (STD). Best results are attained by KCCA (GlvCC, LSA) and are highlighted in boldface.}
\label{tab1}
\end{table}
\begin{table}
\centering
\resizebox{\linewidth}{!}{
\begin{tabular}{|c|@{}c@{}|c|c|c|}
\hline
Data Set& Embedding & Avg Precision & Avg F-score & Avg AUC\\
\hline
Yelp&\begin{tabular}{c}
     GlvCC\\
     \textbf{KCCA(GlvCC, LSA)}\\
     CCA(GlvCC, LSA)\\
     concSVD(GlvCC,LSA)\\
     RNTN
     \end{tabular}&
     \begin{tabular}{c}
     86.47$\pm$1.9\\
     \textbf{91.06$\pm$0.8}\\
     86.26$\pm$1.4\\
     85.53$\pm$2.1\\
     83.11$\pm$1.1
     \end{tabular}&\begin{tabular}{c}
     83.51$\pm$2.6\\
     \textbf{88.66$\pm$2.4}\\
     82.61$\pm$1.1\\
     84.90$\pm$1.7\\
      -
     \end{tabular}&\begin{tabular}{c}
     83.83$\pm$2.2\\
     \textbf{88.76$\pm$2.4}\\
     83.99$\pm$0.8\\
     84.96$\pm$1.5\\
     -
     \end{tabular}\\
\hline
Amazon&\begin{tabular}{c}
     GlvCC\\
     \textbf{KCCA(GlvCC, LSA)}\\
     CCA(GlvCC, LSA)\\
     concSVD(GlvCC, LSA)\\
     RNTN
     \end{tabular}&
     \begin{tabular}{c}
     87.93$\pm$2.7\\
     \textbf{90.56$\pm$2.1}\\
     87.12$\pm$2.6\\
     85.73$\pm$1.9\\
     82.84$\pm$0.6
     \end{tabular}&\begin{tabular}{c}
     82.41$\pm$3.3\\
     \textbf{86.52$\pm$2.0}\\
     83.18$\pm$2.2\\
     85.19$\pm$2.4\\
     -
     \end{tabular}&\begin{tabular}{c}
     83.24$\pm$2.8\\
     \textbf{86.74$\pm$1.9}\\
     83.78$\pm$2.1\\
     85.17$\pm$2.6\\
     -
     \end{tabular}\\
\hline
IMDB&\begin{tabular}{c}
     GlvCC\\
     \textbf{KCCA(GlvCC, LSA)}\\
     CCA(GlvCC, LSA)\\
     concSVD(GlvCC, LSA)\\
     RNTN
     \end{tabular}&
     \begin{tabular}{c}
     54.02$\pm$3.2\\
     59.76$\pm$7.3\\
     53.62$\pm$1.6\\
     52.75$\pm$2.3\\
     \textbf{80.88$\pm$0.7}
     \end{tabular}&\begin{tabular}{c}
     53.03$\pm$5.2\\
     \textbf{53.26$\pm$6.1}\\
     50.62$\pm$5.1\\
     53.05$\pm$6.0\\
     -
     \end{tabular}&\begin{tabular}{c}
     53.01$\pm$2.0\\
     56.46$\pm$3.4\\
     \textbf{58.75$\pm$3.7}\\
     53.54$\pm$2.5\\
     -
     \end{tabular}\\
\hline
A-CHESS&\begin{tabular}{c}
     GlvCC\\
     \textbf{KCCA(GlvCC, LSA)}\\
     CCA(GlvCC, LSA)\\
     concSVD(GlvCC, LSA)\\
     RNTN
     \end{tabular}&
     \begin{tabular}{c}
     52.21$\pm$5.1\\
     \textbf{55.37$\pm$5.5}\\
     54.34$\pm$3.6\\
     40.41$\pm$4.2\\
     -
     \end{tabular}&\begin{tabular}{c}
     \textbf{55.26$\pm$5.6}\\
     50.67$\pm$5.0\\
     48.76$\pm$2.9\\
     44.75$\pm$5.2\\
     -
     \end{tabular}&\begin{tabular}{c}
     \textbf{74.28$\pm$3.6}\\
     69.89$\pm$3.1\\
     68.78$\pm$2.4\\
     68.13$\pm$3.8\\
     -
     \end{tabular}\\
\hline
\end{tabular}
}
\caption{This table shows results obtained by using sentence embeddings from the InferSent encoder in the sentiment classification task. Metrics reported are average Precision, F-score and AUC along with the corresponding standard deviations (STD). Best results are obtained by KCCA (GlvCC, LSA) and are highlighted in boldface.}
\label{tab2}
\end{table}

\section{Experimental Evaluation}\label{exps}
This section evaluates \da~embeddings in binary sentiment classification tasks on four standard data sets. Document embeddings are obtained via (i) a standard framework, i.e document embeddings are a weighted combination of their constituent word embeddings and (ii) by initializing a state of the art sentence encoding algorithm InferSent~\cite{conneau2017supervised} with word embeddings to obtain sentence embeddings. Encoded sentences are then classified using a Logistic Regressor.
\noindent
\subsection{Datasets}
The following balanced and imbalanced data sets are used for experimentation,
\begin{itemize}
\item \textbf{Yelp:} This is a balanced data set consisting of 1000 restaurant reviews obtained from Yelp. Each review is labeled as either `Positive' or `Negative'. There are a total of 2049 distinct word tokens in this data set.
\item \textbf{Amazon:} In this balanced data set there are 1000 product reviews obtained from Amazon. Each product review is labeled either `Positive' or `Negative'. There are a total of 1865 distinct word tokens in this data set.
\item \textbf{IMDB:} This is a balanced data set consisting of 1000 reviews for movies on IMDB. Each movie review is labeled either `Positive' or `Negative'. There are a total of 3075 distinct word tokens in this data set.
\item \textbf{A-CHESS:} This is a proprietary data set\footnote{Center for Health Enhancement System Services at UW-Madison} obtained from a study involving users with alcohol addiction. Text data is obtained from a discussion forum in the A-CHESS mobile app \cite{quanbeck2014mobile}. There are a total of 2500 text messages, with 8$\%$ of the messages indicative of relapse risk. Since this data set is part of a clinical trial, an exact text message cannot be provided as an example. 
However, the following messages illustrate typical messages in this data set, \textit{``I've been clean for about 7 months but even now I still feel like maybe I won't make it.''} Such a message is marked as `threat' by a human moderator. On the other hand there are other benign messages that are marked `not threat' such as \textit{``30 days sober and counting, I feel like I am getting my life back.''} The aim is to eventually automate this process since human moderation involves considerable effort and time. This is an unbalanced data set (~8$\%$ of the messages are marked `threat') with a total of 3400 distinct work tokens.   
\end{itemize}
The first three data sets are obtained from~\cite{kotzias2015group}.

\subsection{Word embeddings and baselines:}
This section briefly describes the various generic and DS embeddings used. We also compare against a basic DA embedding baseline in both the standard framework and while initializing the neural network baseline. 
\begin{itemize}
\item \textbf{Generic word embeddings:} Generic word embeddings used are GloVe\footnote{\url{https://nlp.stanford.edu/projects/glove/}} from both Wikipedia and common crawl and the word2vec (Skip-gram) embeddings\footnote{\url{https://code.google.com/archive/p/word2vec/}}. These generic embeddings will be denoted as Glv, GlvCC and w2v.
\item \textbf{DS word embeddings:} DS embeddings are obtained via Latent Semantic Analysis (LSA) and via retraining word2vec on the test data sets using the implementation in gensim\footnote{\url{https://radimrehurek.com/gensim/}}. DS embeddings via LSA are denoted by LSA and DS embeddings via word2vec are denoted by DSw2v.
\item \textbf{concatenation-SVD baseline:} Generic and DS embeddings are concatenated to form a single embeddings matrix. SVD is performed on this matrix and the resulting singular vectors are projected onto the $d$ largest singular values to form resultant word embeddings. These meta-embeddings proposed by~\cite{yin2016learning} have demonstrated considerable success in intrinsic tasks such as similarities, analogies etc.
\end{itemize}
Details about dimensions of the word embeddings and kernel hyperparameter tuning are found in the supplemental material.

The following neural network baselines are used in this work,
\begin{itemize}
\item \textbf{InferSent:}This is a bidrectional LSTM based sentence encoder~\cite{conneau2017supervised} that learns sentence encodings in a supervised fashion on a natural language inference (NLI) data set. The aim is to use the sentence encoder trained on the NLI data set to learn generic sentence encodings for use in transfer learning applications.
\item \textbf{RNTN:} The Recursive Neural Tensor Network~\cite{socher2013recursive} baseline is a neural network based dependency parser that performs sentiment analysis. Since the data sets considered in our experiments have binary sentiments we compare against this baseline as well.   
\end{itemize}
Note that InferSent is fine-tuned with a combination of GloVe common crawl embeddings and \da~embeddings, and concSVD. The choice of GloVe common crawl embeddings is in keeping with the experimental conditions of the authors of InferSent. Since the data sets at hand do not contain all the tokens required to retrain InferSent, we replace word tokens that are common across our test data sets and InferSent training data with the \da~embeddings and concSVD.

Since we have a combination of balanced and unbalanced test data sets, test metrics reported are Precision, F-score and AUC. We perform 10-fold cross validation to determine hyperparameters and so we report averages of the performance metrics along with the standard deviation.
\section{Results and Discussion}\label{results}
From Tables~\ref{tab1} and~\ref{tab2} we see that \da~embeddings perform better than concSVD as well as the generic and DS word embeddings, when used in a standard classification task as well as when used to initialize a sentence encoding algorithm. As expected, LSA DS embeddings provide better results than word2vec DS embeddings. Note that on the imbalanced A-CHESS data set, on the standard classification task, KCCA embeddings perform better than the other baselines across all three performance metrics. However from Table~\ref{tab2}, GlvCC embeddings achieve a higher average F-score and AUC over KCCA embeddings that obtain the highest precision. 

While one can argue that when evaluating a classifier, the F-score and AUC are better indicators of performance, it is to be noted that A-CHESS is highly imbalanced and precision is calculated on the minor (positive) class that is of most interest. Also note that, InferSent is retrained on the balanced NLI data set that is much larger in size than the A-CHESS test set. Certainly such a training set has more instances of positive samples. Thus when using generic word embeddings to initialize the sentence encoder, which uses the outputs in the classification task, the overall F-score and AUC are better.

From our hypothesis, KCCA embeddings are expected to perform better than the others because CCA/KCCA provides an intuitively better technique to preserve information from both the generic and DS embeddings. On the other hand the concSVD based embeddings do not exploit information in both the generic and DS embeddings. Furthermore, in their work~\cite{yin2016learning} propose to learn an `ensemble' of meta-embeddings by learning weights to combine different generic word embeddings via a simple neural network. We determine the proper weight for combination of DS and generic embeddings in the CCA/KCCA space using the simple optimization problem given in Equation~\eqref{opteq}.

Thus, task specific \da~embeddings formed by a proper weighted combination of DS and generic word embeddings are expected to do better than the concSVD embeddings and individual generic and/or DS embeddings and this is verified empirically.
Also note that the LSA DS embeddings do better than the word2vec DS embeddings. This is expected due to the size of the test sets and the nature of the word2vec algorithm. We expect similar observations when using GloVe DS embeddings owing to the similarities between word2vec and GloVe.
\section{Conclusion}\label{conclusion}
This paper presents a simple yet effective method to learn Domain Adapted word embeddings that generally outperform generic and Domain Specific word embeddings in sentiment classification experiments on a variety of standard data sets. CCA/KCCA based \da~embeddings generally outperform even a concatenation based methods.  
\section*{Acknowledgments}
We would like to thank Ravi Raju for lending computing support for training our neural network baselines. We also thank the anonymous reviewers for their feedback and suggestions.

\bibliography{refbib}
\bibliographystyle{acl_natbib}

\end{document}


\maketitle
\begin{abstract}
{Supplemental material provides details about word tokens, embedding dimensions and hyperparameter details.}
\end{abstract}


\bibliography{refbib}
\bibliographystyle{acl_natbib}


\section{Dimensions of CCA and KCCA projections.} 
Using both KCCA and CCA, generic embeddings and DS embeddings are projected onto their $d$ largest correlated dimensions. By construction, $d \leq \min{(d_{1},d_{2})}$. The best $d$ for each data set is obtained via 10 fold cross validation on the sentiment classification task. Table 2 provides dimensions of all word embeddings considered. Note that for LSA and \da, average word embedding dimension across all four data sets are reported. Generic word embeddings such as GloVe and word2vec are of fixed dimensions across all four data sets.

\section{Kernel parameter estimation.} 
Parameter $\sigma$ of the Gaussian kernel used in KCCA is obtained empirically from the data. The median ($\mu$) of pairwise distances between data points mapped by the kernel function is used to determine $\sigma$. Typically $\sigma= \mu$ or $\sigma = 2\mu$. In this section both values are considered for $\sigma$ and results with the best performing $\sigma$ are reported.
\section{Word tokens and word embeddings dimensions}
\begin{table}
\centering
\resizebox{0.7\columnwidth}{!}{
\begin{tabular}{|c|c|}
\hline
Data Set & Word Tokens\\
\hline
Yelp &2049\\
Amazon & 1865\\
IMDB& 3075\\
A-CHESS &3400\\
\hline
\end{tabular}
}
\caption{This table presents the unique tokens present in each of the four data sets considered in the experiments.}
\end{table}
\begin{table}
\centering
\resizebox{0.8\columnwidth}{!}{
\begin{tabular}{|c|c|}
\hline
Word embedding & Dimension\\
\hline
GloVe & 100\\
word2vec & 300\\
LSA & 70\\
CCA-\da & 68\\
KCCA-\da & 68\\
GloVe common crawl& 300\\
AdaptGloVe& 300\\
\hline
\end{tabular}
}
\caption{This table presents the average dimensions of LSA, generic and \da~word embeddings.}
\label{dim}
\end{table}